\documentclass[a4paper,twoside]{article}

\usepackage{epsfig}
\usepackage{subcaption}
\usepackage{calc}
\usepackage{amssymb}
\usepackage{amstext}
\usepackage{amsmath}
\usepackage{amsthm}
\usepackage{multicol}
\usepackage{pslatex}
\usepackage{apalike}
\usepackage[bottom]{footmisc}
\usepackage{url} % Added Manually
\usepackage{graphicx}
\graphicspath{ {./figs/} }

\usepackage{SCITEPRESS}     % Please add other packages that you may need BEFORE the SCITEPRESS.sty package.

\begin{document}

\title{Instance Segmentation Based Graph Extraction\\for Handwritten Circuit Diagram Images}

\author{\authorname{Johannes Bayer\sup{1}\orcidAuthor{0000-0002-0728-8735}, Amit Kumar Roy\sup{1}\orcidAuthor{0000-0003-4405-752X} and Andreas Dengel\sup{1}\orcidAuthor{0000-0002-6100-8255}}
\affiliation{\sup{1}Deutsches Forschungszentrum f\"ur k\"unstliche Intelligenz,\\ Trippstadter Str. 122, Kaiserslautern, Germany}
\email{\{johannes.bayer, amit.roy, andreas.dengel\}@dfki.de}
}


\abstract{
  Handwritten circuit diagrams from educational scenarios or historic sources usually exist on analogue media. For deriving their functional principles or flaws automatically, they need to be digitized, extracting their electrical graph. Recently, the base technologies for automated pipelines facilitating this process shifted from computer vision to machine learning. This paper describes an approach for extracting both the electrical components (including their terminals and describing texts) as well their interconnections (including junctions and wire hops) by the means of instance segmentation and keypoint extraction. Consequently, the resulting graph extraction process consists of a simple two-step process of model inference and trivial geometric keypoint matching. The dataset itself, its preparation, model training and post-processing are described and publicly available.
}

\onecolumn \maketitle \normalsize \setcounter{footnote}{0} \vfill

\section{\uppercase{Introduction}}
\label{sec:introduction}

Handwritten circuit diagrams still occur nowadays, for example in educational contexts, when communicating swiftly sketched ideas or viewing historic schematics. In most scenarios, automatic means for digitization are desired, i.e. the extraction of electrical graphs from scanned or photographed images for further analysis in computer-aided engineering software.

While the pipelines for graph extraction from engineering diagrams adopted more machine learning over time, they still tend to involve computationally expensive computer vision. At the same time, the provision of datasets for training these models is costly.

The approach described in this paper aims to move more functionality to the machine learning model, allowing for a simplified graph extraction during test time. For both electrical symbols and their interconnections being detected as objects along with their connector points, detailed knowledge on their layout is required during training. The costs for providing the complex training data necessary are mitigated by a modular method.

\section{\uppercase{Related Work}}
\label{sec:related_work}

\subsection{Existing Approaches}

While early approaches to digitize electric circuit diagrams were based on computer vision \cite{bailey1995electronic} and limited to printed diagrams, later works incorporate machine learning like support vector machines \cite{lakshman2019handwritten}, allowing to also process handwritten diagrams. Recently, the trend shifted to deep artificial neural networks \cite{reddy2021hand} \cite{dai2017circuit}. However, all these approaches rely on rather complex and computationally expensive pipelines in which symbols and their connections are processed in individual steps. For example, a dedicated neural network is used for electrical symbol classification \cite{rabbani2016hand}.

Similar trends can also be observed in the closely related domain of piping and instrumentation diagrams for the chemical industry \cite{mani2020automatic} \cite{nurminen2019object} \cite{rahul2019automatic} \cite{sierla2020towards}.

Unfortunately, most literature rely on small datasets, simple circuits, is restricted to special cases like digital logic circuits \cite{majeed2020sketic} or does not use publicly available datasets at all, preventing reproducibility and comparability.

\begin{figure*}[!h]
  \centering
  \includegraphics[width=14.5cm]{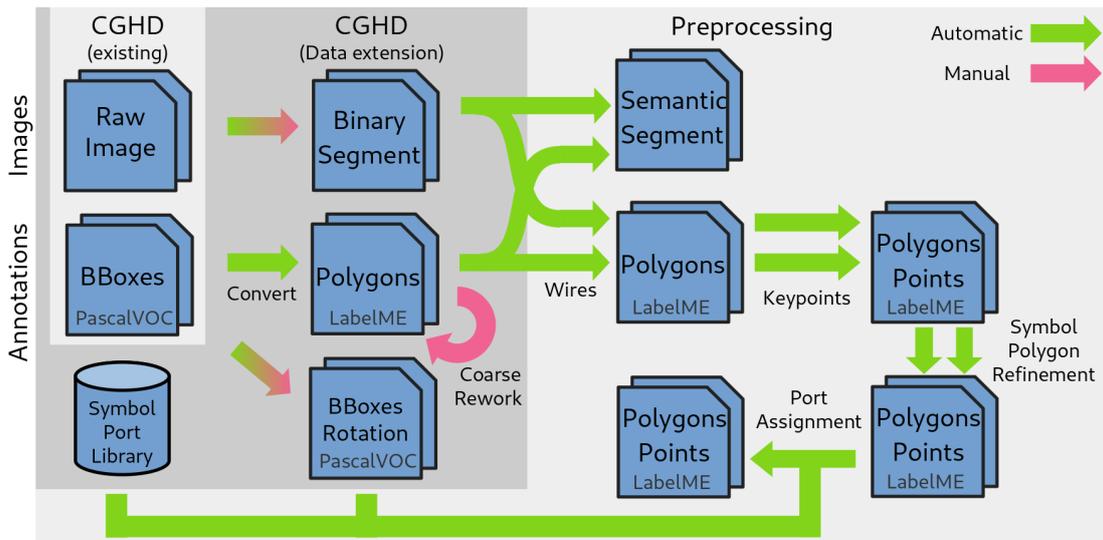}
  \caption{Dataset Preparation Workflow}
  \label{fig:workflow}
\end{figure*}

\subsection{Mask RCNN}

Mask R-CNN \cite{dollar2017mask} is an artificial neural network architecture, allowing for the prediction of bounding boxes, instances masks and keypoints. For the paper at hand, the Detectron 2 \cite{wu2019detectron2} implementation is used.

%marcel2010torchvision

\subsection{CGHD}

CGHD \cite{thoma2021public} is a publicly available dataset of handwritten circuit diagrams with bounding box annotations for the contained electrical symbol, texts (for e.g. component names, electrical properties and circuit headings) and structural elements like junctions and crossovers (wire hops). Since its first description, it has been extended in sample count, the set of classes has also been extended and hierarchically structured. Every circuit is drawn twice, photographed four times and hence occurs as eight samples of pairs of images and annotations in the dataset. This allows for automatically verifying inter-annotator agreement. The individual images are taken under varying conditions of lighting and physical degradation to maximize sample variation. In its current form, it contains $2.208$ raw images with $185.641$ bounding box annotations of $58$ object classes.

\section{\uppercase{Methodology}}
\label{sec:methodology}

The CGHD dataset is extended by instance segmentation ground truth in a set of separated and semi-automated processing steps, which significantly lowers the manual annotation overhead, allows for future reuse and adaptation and avoids annotation ambiguities (see fig.~\ref{fig:workflow}). The extended dataset along with the processing scripts are publicly available \footnote{\url{https://zenodo.org/record/7355865}} \footnote{\url{https://gitlab.com/circuitgraph}}.

A Mask-RCNN model is trained on these extensions to demonstrate the viability of the proposed approach.

The tool set developed in conjunction with this paper allows for graph import and hence links instance segmentation and graph processing.

\subsection{Dataset Preparation}

\begin{figure}[!h]
  \centering
  \includegraphics[width=\linewidth]{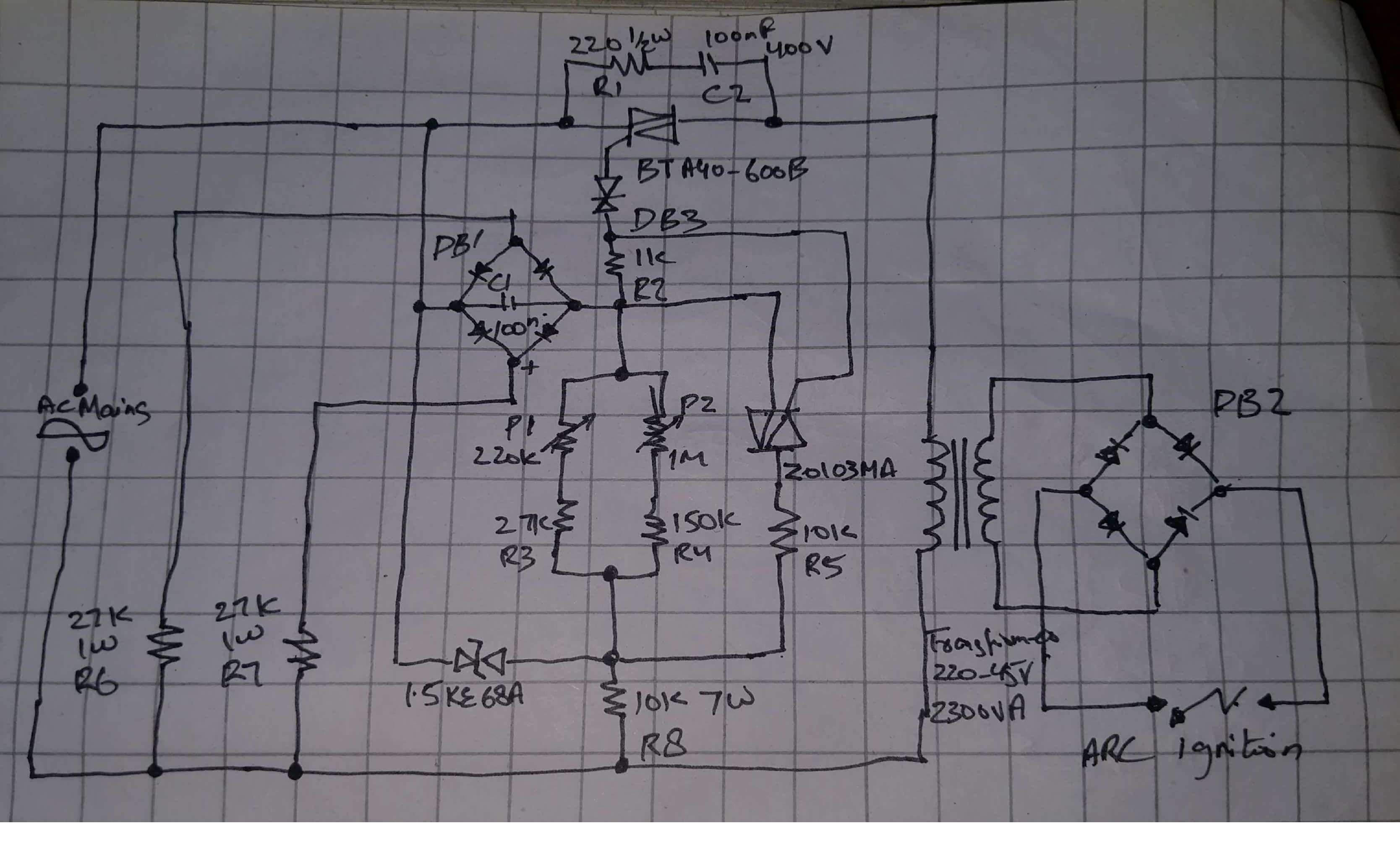}
  \caption{Raw Image Sample from CGHD.}
  \label{fig:sample_raw}
\end{figure}

\begin{figure}[!h]
  \centering
  \includegraphics[width=\linewidth]{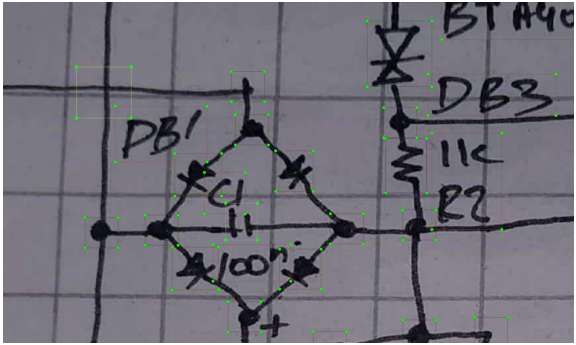}
  \caption{Sample Excerpt showing Overlapping Bounding Box Annotation and Challenging Background.}
  \label{fig:sample_overlap}
\end{figure}

Since the original CGHD dataset (see fig.~\ref{fig:sample_raw}) provides bounding box annotations only for object detection of electrical symbols (and basic structural elements), a graph extraction pipeline built on a respective object detection model requires additional processing for connection extraction. This is computationally expensive and challenged by structured background paper as well as overlaps between bounding box Annotations (see fig.~\ref{fig:sample_overlap}).

\subsubsection{Binary Segmentation Maps}

\begin{figure}[!h]
  \centering
  \includegraphics[width=\linewidth]{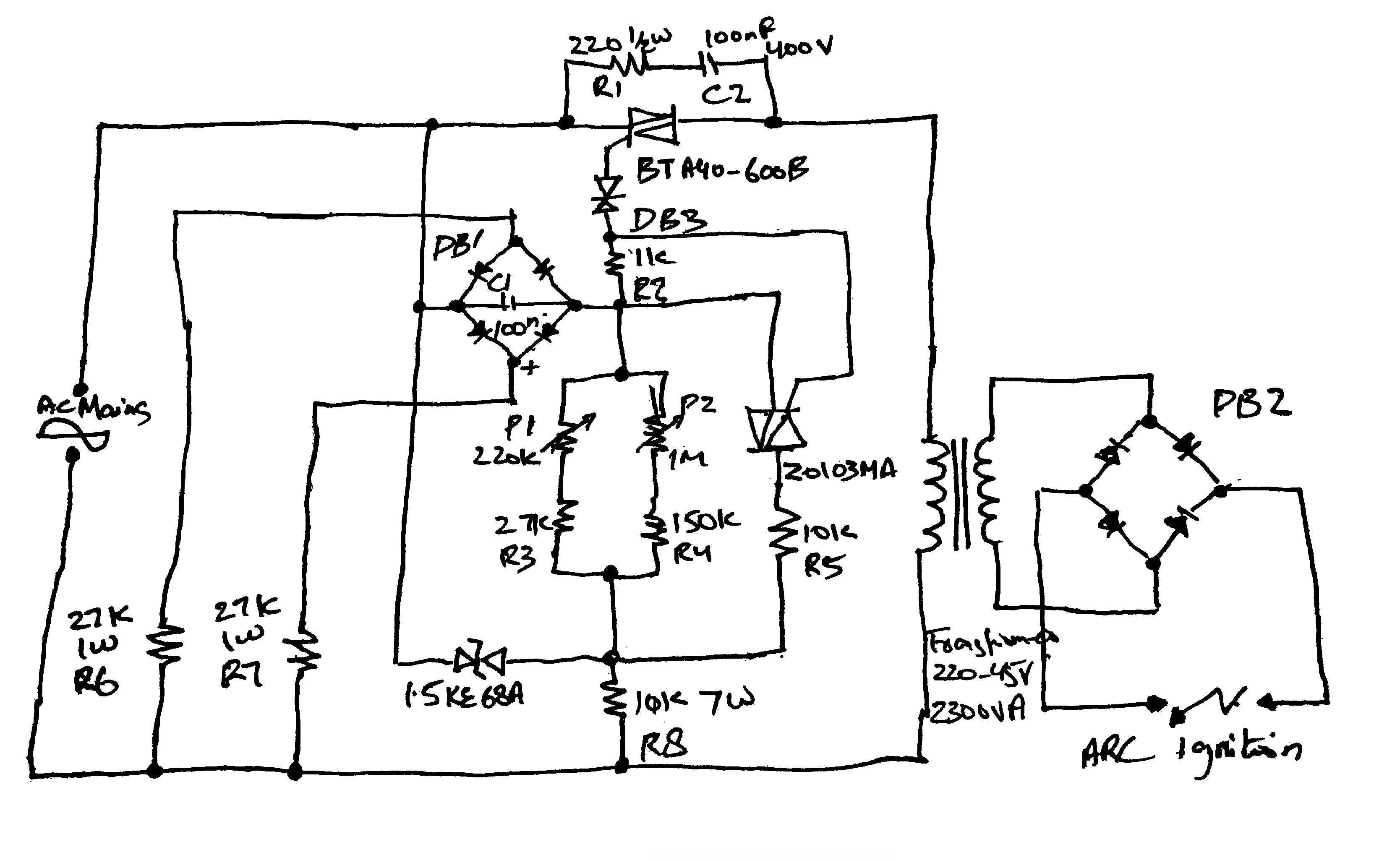}
  \caption{Binary Segmentation Map.}
  \label{fig:example1}
\end{figure}

Through semi-automatic means \cite{gimp} like noise filters, color enhancement, thresholding and manual correction, binary segmentation maps are created from the raw images in which intended drafter's pencil strokes \textit{related to the circuit} are separated from the (e.g. lined or ruled) paper background and surrounding objects. As these maps are stored independently, they can be used for additional segmentation purposes in future.

\subsubsection{Coarse Polygon Masks}

\begin{figure}[!h]
  \centering
  \includegraphics[width=\linewidth]{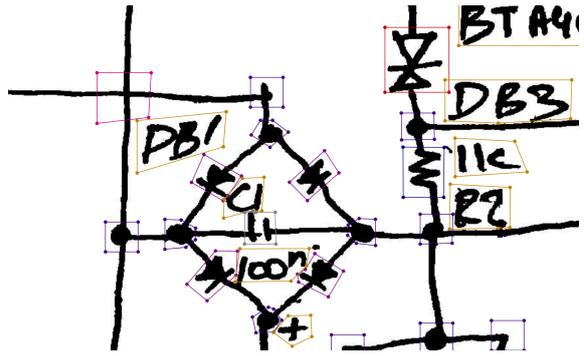}
  \caption{Coarsely Reworked Polygons Avoid Overlaps while Requiring Little Manual Effort.}
  \label{fig:sample_coarse}
\end{figure}

After the bounding box annotations have been converted into polygons, they are manually reworked with respect to avoiding overlaps between individual polygons \textit{in stoke areas} as well as excluding longer parts of connecting wires (see fig.~\ref{fig:sample_coarse}). LabelME \cite{russell2008labelme} is used for this purpose.

\begin{figure}
  \centering
  \includegraphics[width=\linewidth]{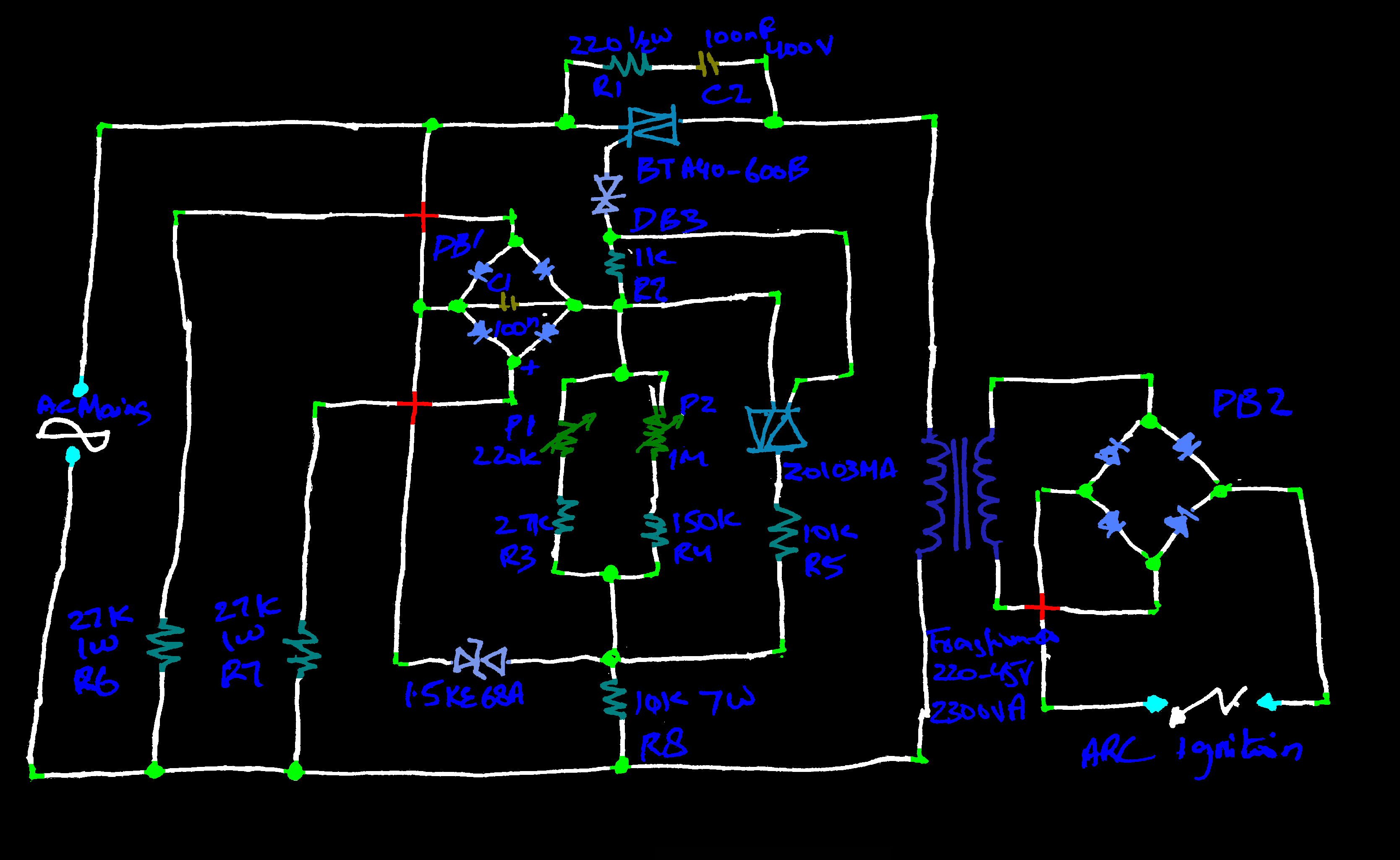}
  \caption{Semantic Segmentation Map.}
  \label{fig:sample_semseg}
\end{figure}

As a mean for simplified visual inspection and to prepare for future semantic segmentation scenarios, a function overlaying coarse polygons with binary segmentation maps is used (see fig.~\ref{fig:sample_semseg}).

\subsubsection{Mask Refinement}

\begin{figure}[!h]
  \centering
  \includegraphics[width=\linewidth]{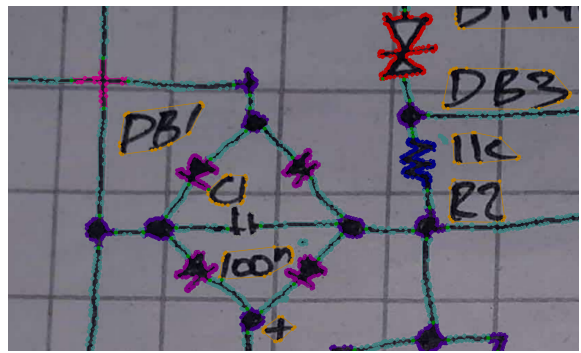}
  \caption{Automatically Refined Polygons.}
  \label{fig:sample_refined}
\end{figure}

Refined polygons are created automatically by applying contour detection to maps obtained by bit-wise \texttt{and} operation between a binary segmentation map and individual coarse polygons. For discontinuous binary map content inside the polygon area, convex hulls are used instead (see fig.~\ref{fig:sample_refined}). As the refined polygons are created algorithmically, they can conveniently be adapted by e.g. altering the polygon sample width. Apart from that, annotation ambiguities are avoided by this step.

\subsubsection{Addition Mask Generation}

All stroke areas not covered by any existing polygon annotation so far is considered an electrical interconnection between components. Hence, wire polygons annotations are created automatically by blacking the polygon annotation areas before applying contour detection, labeling the resulting polygons as \texttt{wire}.

\subsubsection{Keypoint Generation}

Based on the assumption that drafter's strokes touching an electrical symbol's shape are representing the symbol's electrical connections, the keypoints for describing the ports of electrical symbols are derived by calculating intersections between stroke areas in the binary segmentation maps with \textit{borders} of the symbol's shape polygons. In order to avoid false positives, the morphological operation erosion is used before. Additionally, as there many direct stroke connections between electrical symbols and texts (due to inaccurate drawing styles) and texts are assumend not to have electrical connections, areas of text strokes are also removed before the actual keypoint generation.

\subsubsection{Keypoint Port Assignment}

In order to form a viable electrical graph, the extracted keypoints need to be assigned respective electrical terminals. For this purpose, rotation annotations are utilized which represent the angle between a symbol instance and its corresponding \textit{prototype} in the symbol port library. More precisely, a prototype's ports are geometrically transformed to the bounding box of the polygon and before being matched. Since number of ports and their position within the (idealized) prototype is known for the majority of symbol types, an automated verification is performed in this processing step.

\subsection{Post-Processing}

\begin{figure}[!h]
  \centering
  \includegraphics[width=\linewidth]{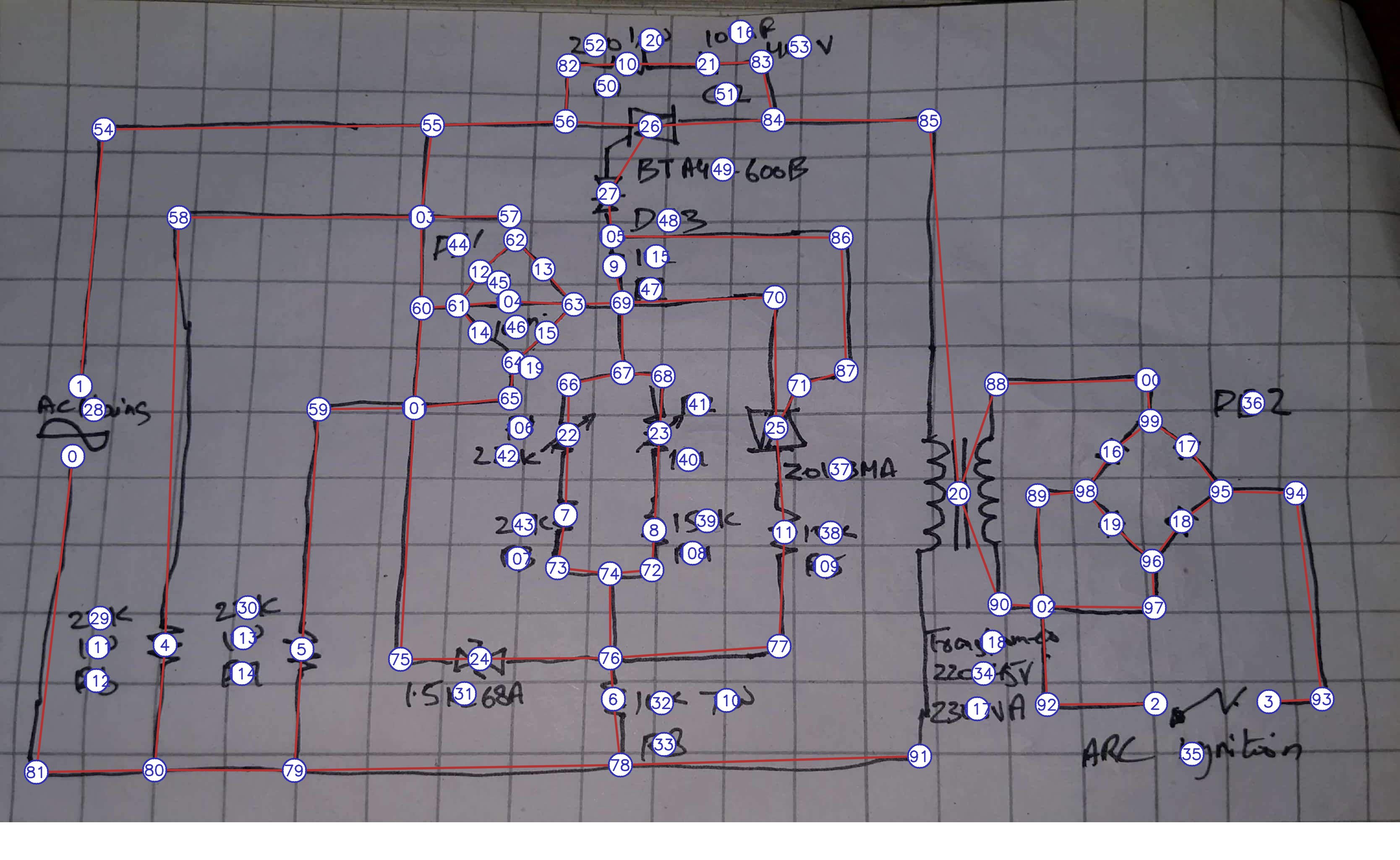}
  \caption{Graph Structure Extracted From Polygon and Keypoint Annotations.}
  \label{fig:sample_graph}
\end{figure}

Constructing electrical graphs from target values or predictions is done by treating all polygons but the \texttt{wire} polygons as nodes, the \texttt{wire} polygons themselves as edges and geometrically matching the keypoints of the \texttt{wire} polygons against all other polygons (see fig.~\ref{fig:sample_graph}).

\section{\uppercase{Experiment}}
\label{sec:experiment}
$245$ Binary segmentation maps with $18.276$ polygon annotations have been created as addition to the existing CGHD dataset.

\begin{figure}[!h]
  \centering
  \includegraphics[width=\linewidth]{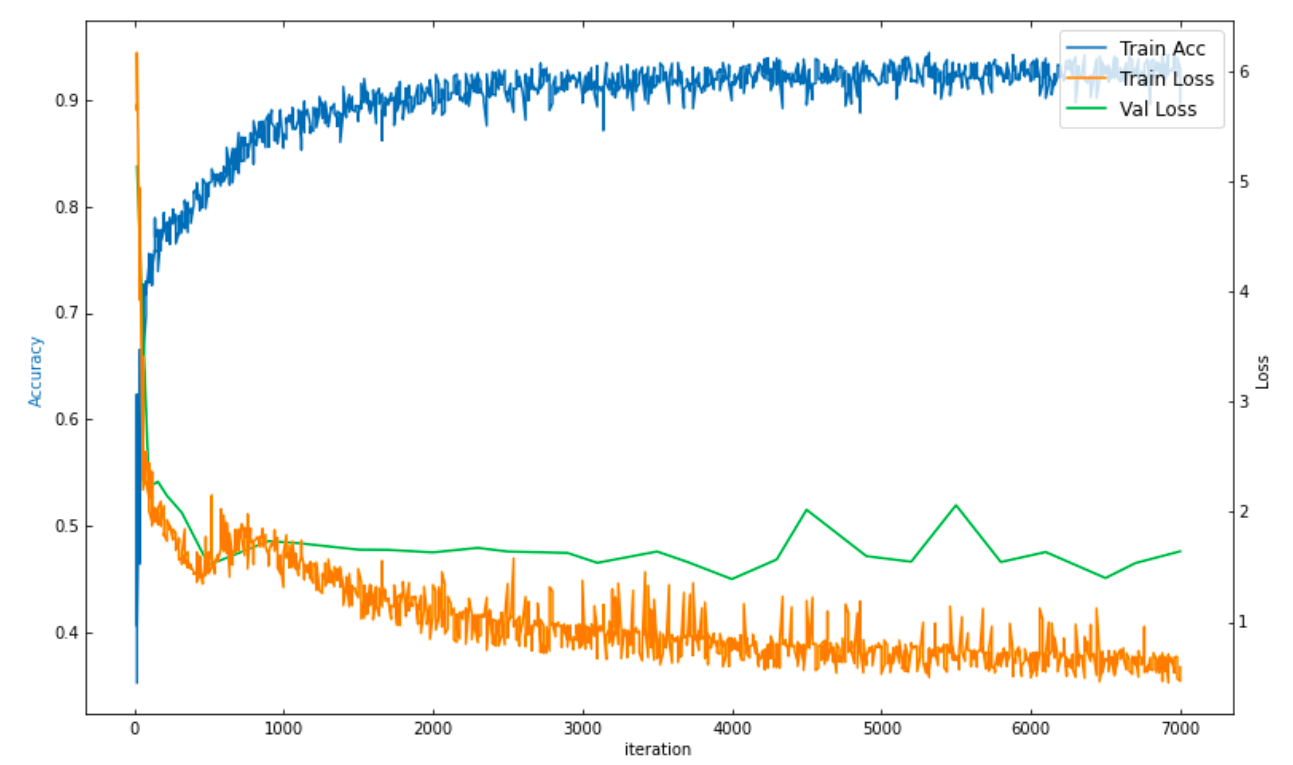}
  \caption{Instance Segmentation Learning Curve.}
  \label{fig:learning_curve}
\end{figure}

\begin{figure}[!h]
  \centering
  \includegraphics[width=\linewidth]{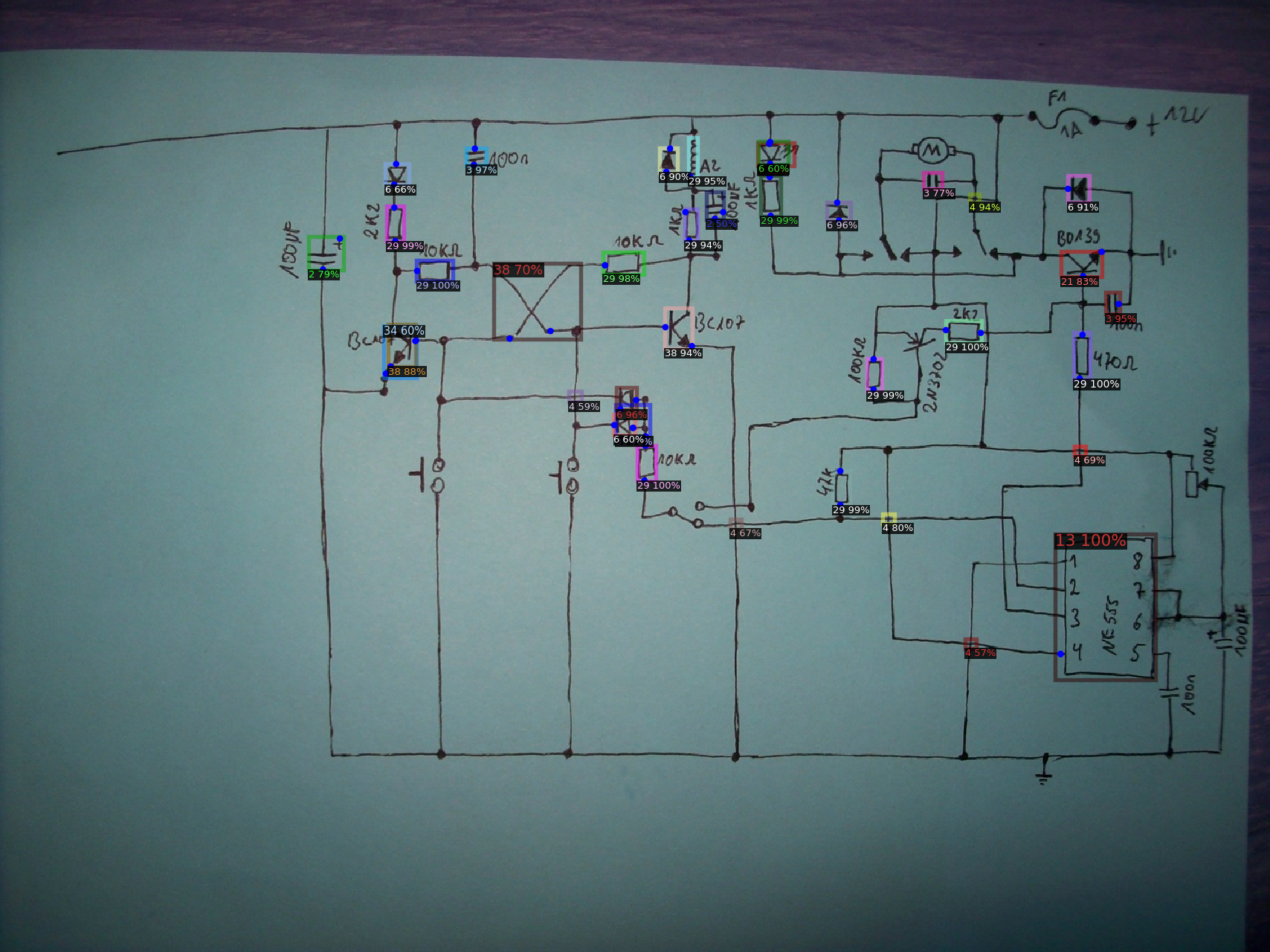}
  \caption{Keypoint Predictions for Electrical Symbols on the Validation Set.}
  \label{fig:keypoint_pred}
\end{figure}

\begin{figure}[!h]
  \centering
  \includegraphics[width=\linewidth]{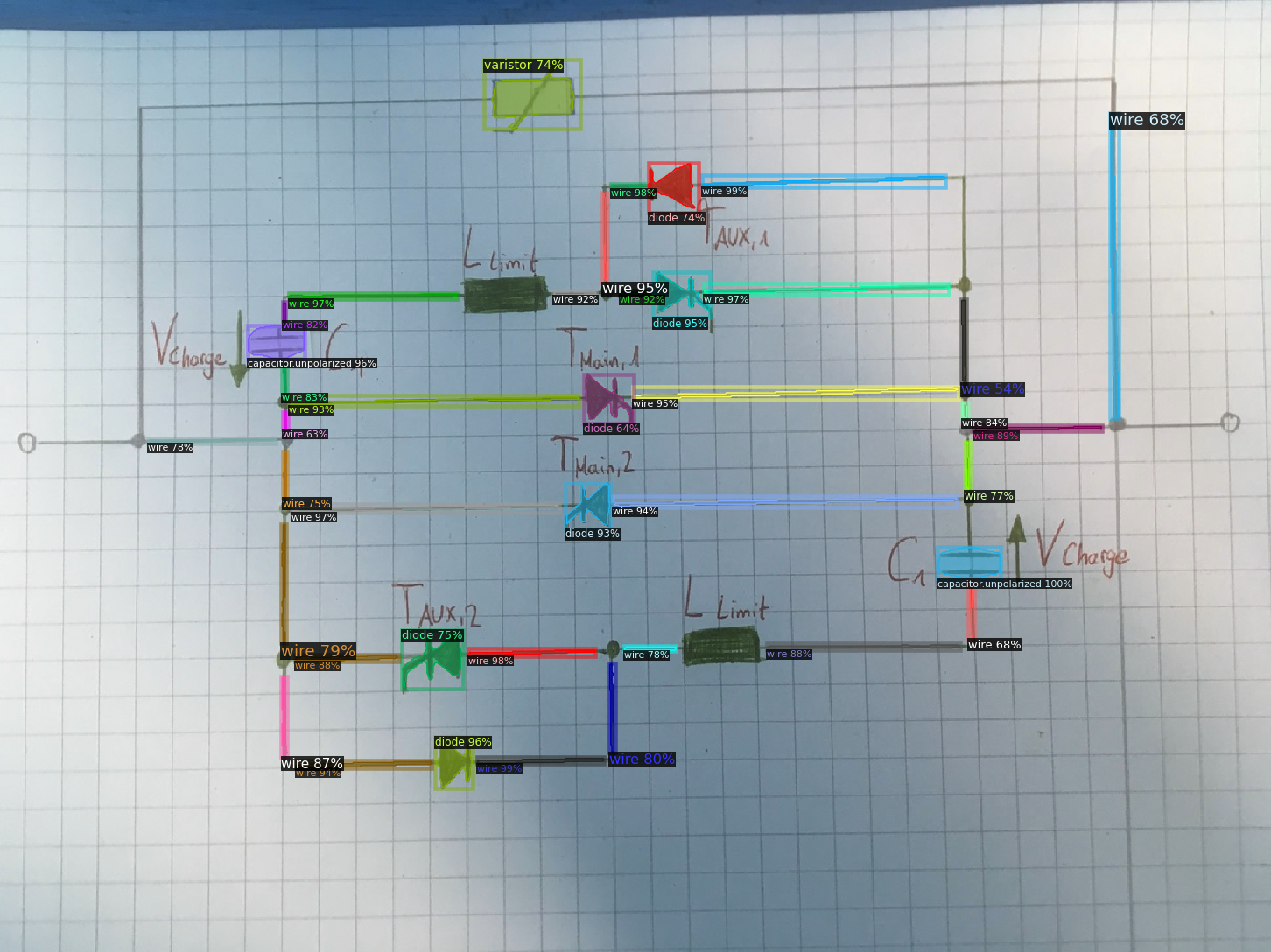}
  \caption{Mask Predictions on Validation Set (junctions and texts omitted).}
  \label{fig:mask_pred}
\end{figure}

A Mask RCNN trained with learning rate of $0.0005$ and a batch size of $4$ for $7000$ iterations resulted in a minimum training loss of $0.44$, a maximum training mask accuracy of $0.94$ and a minimum validation loss of $1.39$ (see fig.~ \ref{fig:learning_curve}). A visual inspection on validation set mask (see fig.\ref{fig:mask_pred}) and keypoint (see fig.~\ref{fig:keypoint_pred}) predictions shows reasonable, yet incomplete recognition.

\section{\uppercase{Conclusions}}
\label{sec:conclusion}

The semantic and instance segmentation additons to the CGHD dataset establish a new, flexible standard for further research on graph extraction from hand-drawn schematics, allowing for arbitrary new methods to be evaluated.

While the experiment demonstrated the general viability of the approach, further model optimizations are required to achieve error-free graph reconstruction.

\section{\uppercase{Future Work}}
\label{sec:future_work}

So far, the described approach is limited in various ways: Most importantly, the types of individual component connectors are not differentiated, which is critical for an accurate simulation of non-linear electrical circuits. Adding a rotation prediction head to the Mask RCNN as well as providing these information in the dataset (which can in turn be semi-automated by a classic template matching) can be used in conjunction with a component library to identify these connector types. Furthermore, drawing errors like discontinuous wires need to be mitigated, which could be done by post-processing with graph neural networks. Apart from that, OCR information need to be incorporated to predict not only the position but also the content of the text labels. Additionally, edge types different from electrical connections need to be identified like mechanical coupling of switches or inductive coil coupling in complex transformers. Finally, as mask information could only be provided for a subset of the original dataset, a join training with full dataset on both masks and bounding boxes only need to be considered.

%\begin{table}[h]
%\vspace{-0.2cm}
%\caption{This caption has more than one line so it has to be
%justified.}\label{tab:example2} \centering
%\begin{tabular}{|c|c|}
%  \hline
%  Example column 1 & Example column 2 \\
%  \hline
%  Example text 1 & Example text 2 \\
%  \hline
%\end{tabular}
%\end{table}

\section*{\uppercase{Acknowledgements}}

The authors cordially tank all drafters and annotators for contributing to the dataset. The reseach for this paper was partly funded by the BMWE (Bundesministerium für Wirtschaft und Klimaschutz), project ecoKI, funding number: 03EN2047B.

\bibliographystyle{apalike}
{\small
\bibliography{paper}}

\end{document}